\documentclass{article}
\usepackage{microtype}
\usepackage{graphicx}
\usepackage{subfigure}
\usepackage{booktabs} 

\usepackage{hyperref}

\usepackage[accepted]{icml2025} 

\usepackage{amsmath}
\usepackage{amssymb}
\usepackage{mathtools}
\usepackage{amsthm}
\usepackage{verbatim} 
\usepackage{subcaption}
\usepackage{pifont}
\usepackage{array}           
\definecolor{customgreen}{RGB}{109, 192, 93}
\usepackage{caption} 
\usepackage{colortbl} 

\usepackage[capitalize,noabbrev]{cleveref}

\theoremstyle{plain}

\theoremstyle{definition}

\theoremstyle{remark}

\usepackage[textsize=tiny]{todonotes}

\icmltitlerunning{Leveraging LLMs as Meta-Judges: 
A Multi-Agent Framework for Evaluating LLM Judgments}

\begin{document}

\twocolumn[
\icmltitle{Leveraging LLMs as Meta-Judges: \\
A Multi-Agent Framework for Evaluating LLM Judgments}



\icmlsetsymbol{equal}{*}

\begin{icmlauthorlist}
\icmlauthor{Yuran Li}{mcgill}
\icmlauthor{Jama Hussein Mohamud}{mila}
\icmlauthor{Chongren Sun}{mcgill}
\icmlauthor{Di Wu}{mcgill}
\icmlauthor{Benoit Boulet}{mcgill}
\end{icmlauthorlist}

\icmlaffiliation{mcgill}{Intelligent Automation Lab, McGill University}
\icmlaffiliation{mila}{Mila, Quebec AI Institute}

\icmlcorrespondingauthor{Yuran Li}{yuran.li@mail.mcgill.ca}

\vskip 0.3in
]


\printAffiliationsAndNotice{}  

\begin{abstract}
 Large language models (LLMs) are being widely applied across various fields, but as tasks become more complex, evaluating their responses is increasingly challenging. Compared to human evaluators, the use of LLMs to support performance evaluation offers a more efficient alternative. However, most studies focus mainly on aligning LLMs' judgments with human preferences, overlooking the existence of biases and mistakes in human judgment. Furthermore, how to select suitable LLM judgments given multiple potential LLM responses remains underexplored. To address these two aforementioned issues, we propose a three-stage meta-judge selection pipeline: 1) developing a comprehensive rubric with GPT-4 and human experts, 2) using three advanced LLM agents to score judgments, and 3) applying a threshold to filter out low-scoring judgments. Compared to methods using a single LLM as both judge and meta-judge, our pipeline introduces multi-agent collaboration and a more comprehensive rubric. Experimental results on the JudgeBench dataset show about 15.55\% improvement compared to raw judgments and about 8.37\% improvement over the single-agent baseline. Our work demonstrates the potential of LLMs as meta-judges and lays the foundation for future research on constructing preference datasets for LLM-as-a-judge reinforcement learning.
\end{abstract}

\section{Introduction}
With the growing adoption of Large Language Models (LLMs) across various domains and applications, particularly in fields such as medicine, law, and education, ensuring the quality of their responses has become increasingly critical\cite{hadi2023survey}. Evaluating the quality of LLMs' responses help us discard incorrect information and provide feedback to improve the quality of LLMs' responses~\cite{chang2024survey}.

The existing evaluation frameworks for the quality of LLMs' output include human evaluation, automated bench-marking and model-based evaluation. Human evaluation is costly, time-consuming, and limited by the bias and capability of human evaluators. Automated benchmarking is constrained by its fixed form and also requires human annotation. In contrast, model-based evaluation overcomes these limitations, accommodating diverse input formats while being cost-effective and time-efficient.
Therefore, recently, numerous works have emerged exploring how LLMs act as evaluators of LLMs' output. These proposals vary in their approaches to configuring LLMs as judges, ranging from simple prompting techniques (zero-shot, few-shot, chain-of-thought) \cite{15-stahl2024exploring} to training LLMs on sample judgment data \cite{16-chan2023chateval}. The approaches also differ in their use of LLMs (single judge versus multiple judges \cite{12-verga2024replacing}) and the level of engagement of the judges (single-turn prompts versus debates \cite{17-khan2024debating}). 

Meanwhile, just as we need to evaluate the quality of LLMs' responses, we also need to assess the quality of their judgments. Previous research has focused on evaluating LLMs' judgments based on alignment with human preference\cite{10-zheng2023judging}. However, in factually and logically complex tasks that exceed human capabilities, LLMs have demonstrated significant advantages over human\cite{tan2024judgebench}. In such cases, using human alignment as the primary evaluation metric becomes less applicable. An automated LLM-meta-judge pipeline for evaluating judgment data could enhance LLMs' ability to address superhuman tasks. As a result, recent research has been actively exploring advancements in LLM-as-a-judge through self-reflection, enabling models to evaluate their judgments (LLM-as-meta-judge) \cite{14-wu2024meta} \cite{trivedi2024self}. These studies show significant potential for improving LLMs' judgment capabilities through meta-judging. However, current methods only adopt simple meta-judging methods, using the same model for both judging and meta-judging tasks. Furthermore, the performance of meta-judging is evaluated indirectly, based solely on its impact on LLM-as-a-judge results, without directly assessing the precision or accuracy of meta-judging itself.

A systematic approach to meta-judging remains largely unexplored. To address this significant research gap, we propose a meta-judging approach that incorporates comprehensive rubric design, a multi-agent module, and score-based threshold selection. Moreover, we benchmark the performance of various LLMs as meta-judges in different configurations.

Our main contributions are:
\begin{itemize}
    \item We establish a complete meta-judging pipeline incorporating a multi-agent module to accurately identify and select correct judgments from a mixed set.

\item We extensively evaluate diverse large language models with various prompt configurations and multi-agent strategies, demonstrating significant precision improvement and analyzing underlying factors.

\end{itemize}


\section{Related Work}

\subsection{Evaluation of LLM Responses} 
As LLMs continue to advance, evaluating the quality of their responses has become a critical research focus. Guo \cite{1-guo2023evaluating} highlights key evaluation aspects, including knowledge and capability, alignment, and safety. Evaluation methods typically fall into three categories: automated benchmarking, human evaluation, and model-based evaluation, each with distinct advantages and limitations \cite{2-huggingface2023llmevaluation}.
Automated benchmarking uses datasets (e.g., TruthfulQA \cite{3-lin2021truthfulqa}, BIG-bench \cite{4-srivastava2022beyond}, and SelfAware \cite{5-yin2023large}) and metrics to evaluate LLMs on specific tasks \cite{zhang2022mme} \cite{zhang2019bertscore} but faces issues like dataset contamination, difficulty in task decomposition, and normalization challenges. Human evaluation allows for open-ended assessments and avoids contamination but is costly and prone to biases, such as reliance on first impressions and domain knowledge gaps \cite{13-wang2023aligning}.
Model-based evaluation provides a more efficient alternative, using high-capability generalist models or specialized models trained on preference data. Methods like NLI-based approaches \cite{6-maynez2020faithfulness}, QAQG \cite{7-manakul2023mqag}, FActScore \cite{8-min2023factscore}, and self-evaluation \cite{9-manakul2023selfcheckgpt} are prominent but suffer from biases, scoring inconsistencies, and occasional misalignment with human rankings. Techniques encouraging reasoning before scoring can reduce these issues, though challenges in consistency and interpretability persist.

\subsection{Evaluation of LLM-as-a-Judge} 
The LLM-as-a-judge paradigm offers a scalable alternative to the human evaluation of LLMs. Zheng \cite{10-zheng2023judging} explored strong LLMs as judges for open-ended questions, identifying limitations such as positional bias, verbosity, self-enhancement bias, and limited reasoning while proposing mitigation strategies and verifying alignment with human preferences. Thakur \cite{11-thakur2024judging}, using the TriviaQA benchmark, emphasized Cohen’s kappa over percent agreement for evaluating alignment, revealing that models aligned with human judgments may not excel at ranking tasks. Verga \cite{12-verga2024replacing} demonstrated that Panels of LLM Evaluators (PoLL) composed of smaller, disjoint model families outperform single large models by reducing intra-model bias at significantly lower costs.
However, these studies rely solely on alignment with human preferences as the performance metric, overlooking human errors and biases, as well as tasks beyond human evaluators' capabilities. Additionally, they do not propose effective methods for selecting correct judgments.

\begin{figure}[htp]
    \centering
    \includegraphics[width=0.98\linewidth]{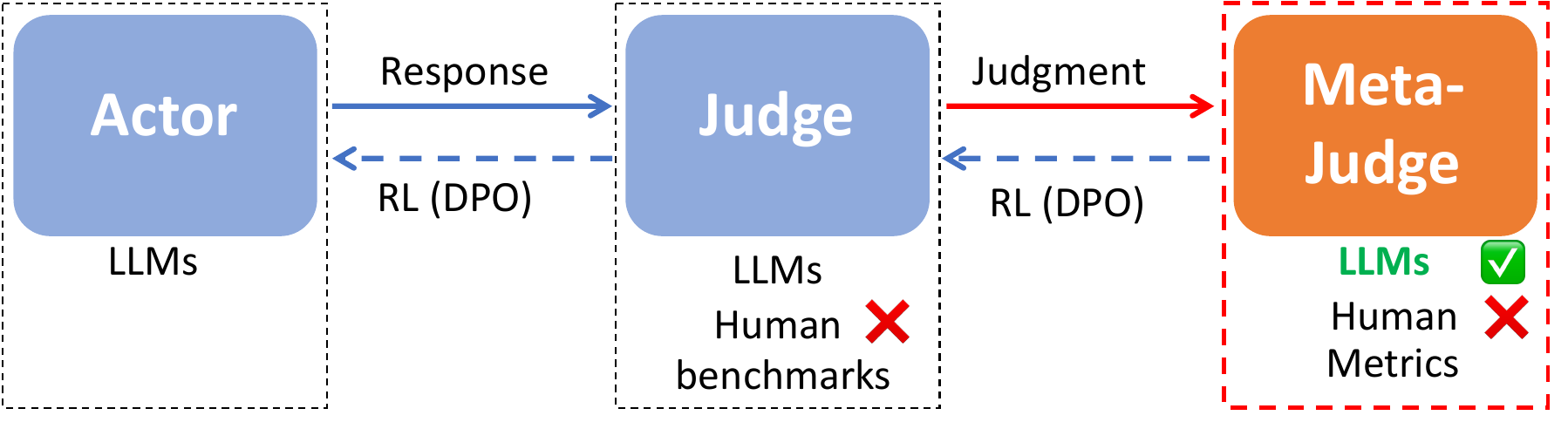} 
    \caption{Method components and interactions. The actor (an LLM) generates responses, the judge evaluates their quality, and the meta-judge assesses the judge’s evaluation. In the diagram, the backward RL (DPO) arrow represents Reinforcement Learning from AI Feedback (RLAIF), using DPO as the training method.}
    \label{fig:overall diagram} 
\end{figure}

Currently, some works have been done to bypass human annotation, and evaluate LLM-as-a-judge in a more objective perspective.
As in the paper, JudgeBench \cite{tan2024judgebench}, Tan argues that human preferences are unreliable indicators of factual accuracy and logic in complex tasks. They propose a pipeline to generate challenging response pairs for LLMs to assess the precision of their judgments against objective ground truth labels, eliminating reliance on human annotation. 
Meanwhile, some experts have proposed that a promising way for improving LLM-as-judge is self-reflection on their own judgments (meta-judge), as discussed in \cite{14-wu2024meta} and \cite{trivedi2024self}. This approach leverages the meta-judge capabilities of LLMs to refine and enhance their own evaluation processes. 
These studies highlight a critical aspect of evaluating the judgment capabilities of LLMs through meta-judging and underscore the importance of accurately selecting correct judgments generated by them. The information flow for responses evaluation, judgments evaluation, and reinforcement learning from AI feedback is illustrated in Figure \ref{fig:overall diagram}.

\section{Approach}
\begin{figure*}[h]
    \centering
    \includegraphics[width=0.9\linewidth]{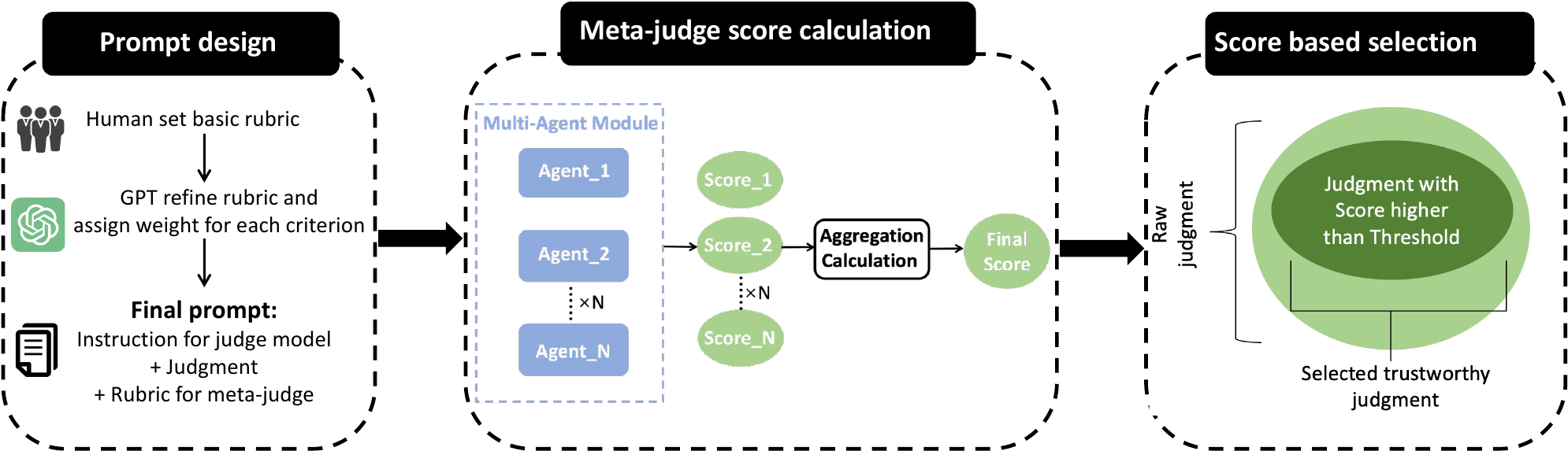} 
    \caption{LLM-as-meta-judge framework. The rubric is predefined in the prompt design stage. We benchmark the judgment using N agents, each providing a score based on a rubric. These N scores are then aggregated through metric calculations to yield a comprehensive score reflecting the LLM judge's performance.}
    \label{fig:meta-judge} 
\end{figure*}

In this section, we introduce a three-stage approach for effectively selecting correct judgments from the raw judgment dataset generated by LLMs. the framework is indicated in Figure~\ref{fig:meta-judge}. Firstly, at the prompt design stage, GPT-4 improves the basic human-crafted rubric into detailed descriptions and assigns weight to each criterion. In meta-judge score calculation stage, $N$ LLMs collaborate to score the judgements based on the preset rubric. In the selection stage, the meta-judge score shows the judgments' alignment with the ground truth label and the judgments with a meta-judge score higher than the threshold are selected as trustworthy judgments.
\subsection{Prompt Design}
As illustrated in Figure \ref{fig:meta-judge}, the Meta-Judge framework begins with the prompt design stage, where human input is initially required to define a basic rubric. GPT-4 is then utilized to refine the rubric into a detailed description and explain the corresponding scoring system.
The rubric consists of seven criteria on a scale of 1 to 5, which collectively evaluate the judgment from multiple perspectives, including accuracy, logical soundness, completeness of evaluation, fairness, relevance to context, clarity of explanation, and impactfulness. An example of a criterion is shown in Table~\ref{tab:rubric-describe0}.
Furthermore, GPT-4 can assign weights to each criterion based on the specific meta-judge scenario. 

The final input prompt includes instruction for the LLM-as-a-judge, the judgment (comprising a decisive conclusion and explanation), and the rubric for the meta-judge. 

\begin{table}[h!]
    \centering
    \setlength{\tabcolsep}{8pt} 
    \renewcommand{\arraystretch}{1.5} 
    \begin{tabular}{p{0.95\columnwidth}}
        \toprule
       \rowcolor{gray!20} \textbf{Criterion:} \textit{Logical Soundness} \\
        \midrule 
         \textbf{Description:} \textit{Measures if the decision-making process follows a clear and logical reasoning path.} \\
        \midrule
        \rowcolor{customgreen!10} \textbf{Score 1:} \textit{Illogical, with no clear reasoning or consistency.}\\ 
        \rowcolor{customgreen!20} \textbf{Score 2:} \textit{Mostly illogical, with several gaps or flawed reasoning.}\\ 
        \rowcolor{customgreen!30} \textbf{Score 3:} \textit{Partially logical, with some inconsistencies in the reasoning process.}\\ 
        \rowcolor{customgreen!50} \textbf{Score 4:} \textit{Mostly logical, with minor flaws in reasoning.}\\ 
        \rowcolor{customgreen!70} \textbf{Score 5:} \textit{Highly logical, with a clear and consistent reasoning process throughout.}\\ 
        \bottomrule
    \end{tabular}
    \caption{Example of meta-judging rubric. This is a short version. In the experiment section, a longer version is also adopted, featuring a three-sentence description and a two-sentence explanation of the scoring system.}
    \label{tab:rubric-describe0}
\end{table}

\subsection{Meta-Judge}
In the meta-judge stage, a multi-agent module comprising N advanced LLMs evaluates the judgments of the LLMs based on the rubric established during the initial prompt design phase. 
Each agent can generate its meta-judge score for each criterion, either independently or collaboratively through discussion. The scores from each agent for each criterion are then aggregated to produce a comprehensive final score.

Three multi-agent collaboration strategies are put forward: weighted averaging, majority voting, and panel discussion.

\subsubsection{Weighted averaging}
When we adopt the weighted average strategy, the meta-judge score is independently generated by each agent, then the final score is calculated as equation (\ref{eq:weight_and_sum}): 
\begin{equation}
    \text{final\_score} = \sum\limits_{i=1}^{M}w_{i}^{agent}\sum\limits_{j=1}^{N}{w_j^{c}}*S_{ij}
    \label{eq:weight_and_sum}
\end{equation}
where $S_{ij}$ represents the score generated by agent $i$ for criterion $j$.
The weight $w^c_j$ reflects the varying importance of different criterion, $\sum_{j=1}^{N}{w_j^{c}}=1$. 
${w_{i}^{agent}}$ denotes the weight of the agent $i$.
For each task $\sum_{i=1}^{M}{w_{i}^{agent}}=1$. 
Both weights can be dynamically adjusted to suit specific downstream tasks.

\subsubsection{Majority Voting}
When majority voting is adopted, each agent independently generates a score. The number of scores that exceed the selection threshold is counted. If more than half of the agents assign scores above the threshold, the final score is set to 5; otherwise, it is set to 1. It is calculated as equation (\ref{eq:majority_voting}):
\begin{equation}
    \text{final\_score} =
    \begin{cases}
        5, & \text{if } \sum_{i=1}^{M} \mathbb{I}(S_i > T) > \frac{M}{2}, \\
        1, & \text{otherwise}.
    \end{cases}
    \label{eq:majority_voting}
\end{equation}
where $S_i = \sum\limits_{j=1}^{N}{w_j^{c}}*S_{ij}$, is the score generated by a single agent. $\mathbb{I}(S_i > T)$ is an indicator function, it equals 1, if $S_i > T$, and $0$ otherwise.

\subsubsection{Panel discussion}
As shown in papers \cite{li2024matevalmultiagentdiscussionframework} and \cite{chan2023chateval}, having each agent play a distinct role and engage in collaborative discussions can lead to less biased and higher-quality outputs.
Therefore, in the multi-agents module, collaborative reasoning is designed to derive the meta-judge score through a collaborative discussion among the agents.

\begin{figure}[htbp]
    \centering
    \includegraphics[width=\linewidth]{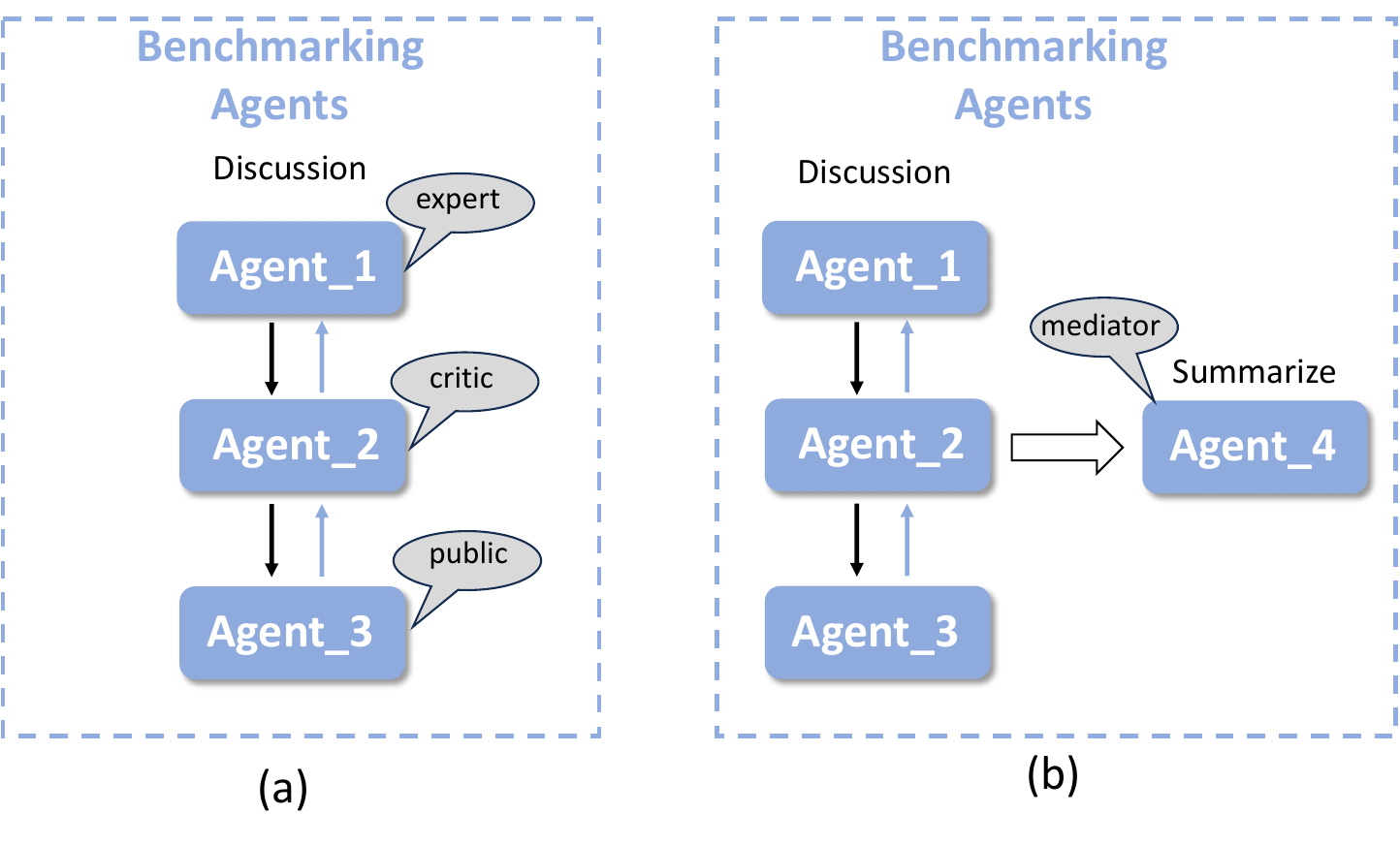} 
    \caption{Cooperative discussion diagram. Each agent is assigned a distinct role. Agent 1 shares its meta-judgment with Agent 2, which updates its own judgment accordingly. Agent 3 then refines its meta-judgment by integrating inputs from Agents 1 and 2. An additional Agent 4 can be applied to summarize all meta-judgments to produce the final outcome.}
    \label{fig:discussion} 
\end{figure}

To facilitate comprehensive learning among agents, we developed the collaborative discussion diagram shown in Fig.~\ref{fig:discussion}. 
When no summarization agent, the final meta-judge score is calculated using equation (\ref{eq:majority_voting}) or equation (\ref{eq:weight_and_sum}). It is shown in Fig.~\ref{fig:discussion}(a).
When a summarization agent is applied, the final score calculation can be simplified into Equation (\ref{eq:weight_and_sum_simple}).
\begin{equation}
    \text{final\_score} = \sum\limits_{j=1}^{N}{w_j^{c}}*S_{j}
    \label{eq:weight_and_sum_simple}
\end{equation}

\subsection{Selection Based on Meta-Judge Score}
According to the preset rubric, the final meta-judge score indicates the reliability of judgment.
Therefore, the meta-judge score can be used to filter out those unreliable judgments and only keep those high-quality ones that provide more reliable, decisive conclusions and more helpful explanations for their judgments. 

To evaluate the capability of LLMs as meta-judges, it is necessary to verify whether judgments with high scores are truly correct. Specifically, we must determine a score threshold above which judgments can be considered reliable.
Hence, a threshold $T$ is set, then judgments with a meta-judge score higher than the threshold are selected as correct judgments, while the judgment with a meta-judge score lower than the threshold will be discarded. This threshold-based selection process enables the evaluation of precision, which serves as an indicator of the LLMs' effectiveness as meta-judges.

\section{Experiment}

\subsection{Experiment Setup}
\textbf{Datasets:}
We select JudgeBench\cite{tan2024judgebench} as source dataset to evaluate how well our method performs. JudgeBench combines challenging datasets such as MMLU-Pro\cite{wang2024mmlu}, LiveBench\cite{white2024livebench}, and LiveCodeBench\cite{jain2024livecodebench}, comprising 154 questions on knowledge, 98 on reasoning, 56 on mathematics, and 42 on coding, to generate difficult-to-distinguish response pairs for LLM evaluation. It includes 350 unique response pairs generated by GPT-4o and 270 unique response pairs generated by Claude-3.5-Sonnet. The datasets contain objective ground truth labels and algorithms to verify correctness. 

\begin{figure}[htbp]
    \centering
    \includegraphics[width=0.6\linewidth]{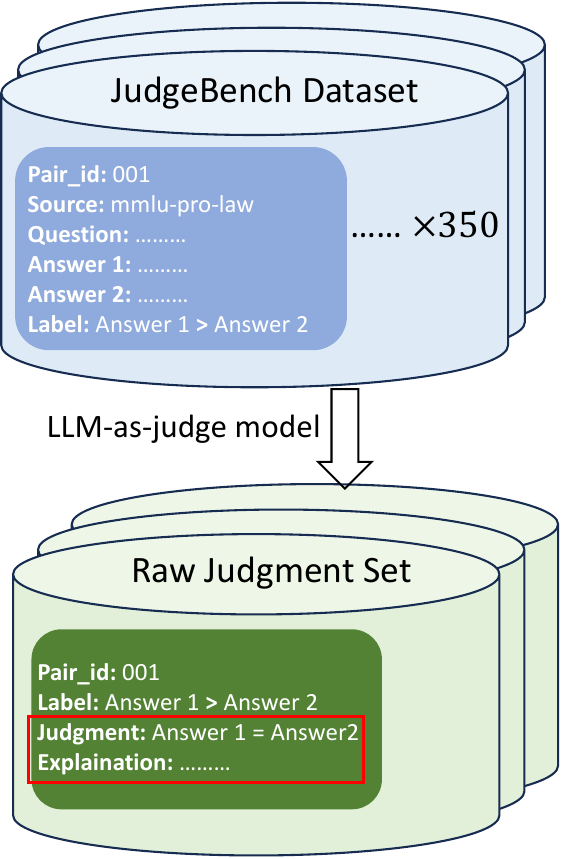} 
    \caption{Generation of Raw Judgments from the JudgeBench Dataset. A judgment is considered true if it matches the label for the answer pairs; otherwise, it is false.}
    \label{fig:dataset} 
\end{figure}

On the basis of the JudgeBench dataset, multiple judgments for response pairs could be generated. The collection of these judgments form our meta-judge experimental dataset, referred to as the raw judgment dataset.
An example of a raw judgment dataset is shown in Fig.~\ref{fig:dataset}.

\paragraph{Evaluation Metric:}
JudgeBench provides the ground truth labels for the response pairs. The judgment GT label is determined based on whether it matches the ground truth label of the corresponding response pair.
we compute precision between the GT labels and our generated labels as equation (\ref{eq:precision_1}):

\begin{equation} \text{Precision} = \frac{N_{correct}}{N_{selected}} = \frac{TP}{TP + FP} \label{eq:precision_1} \end{equation} 

The reason we choose precision as our primary evaluation metric is that, for effective meta-judging to improve judging performance, providing high-quality and correct meta-judgments is our top priority. Minimizing false positives, where incorrect meta-judgments are classified as correct, is our key objective.

\paragraph{Benchmarking Agents:}
We choose four advanced LLMs as our benchmarking agents: GPT-4o, GPT-4o-mini, Claude-3.5-Sonnet, and LLaMa-3.1-405B-Instruct.

\paragraph{Baselines and Experimental Configurations:}
To date, meta-judging has been explored only in single-agent settings. Therefore, for comparison, multiple meta-judge agents are first evaluated individually to identify the best-performing single-agent strategy. Then, different collaboration approaches in multi-agent setups are compared. 

For the single-agent meta-judge comparison, we consider the following configurations:
(1) Baseline meta-judge method as in paper \cite{14-wu2024meta}.
(2) Short Scoring Rubric: Meta-judging with a short rubric to generate a score (soft label).
(3) Long Scoring Rubric: Meta-judging with a long rubric to generate a score (soft label).
(4) Binary Rubric: Meta-judging to generate a True/False decision (hard label).

For the multi-agent meta-judge comparison, we consider the following approaches:
(1) Weighted Averaging: calculates the meta-judge score by taking a weighted average of the scores independently derived by each agent.
(2) Majority voting: Determines the meta-judge score by counting how many agents' scores exceed a predefined threshold.
(3) Panel discussion: derives the meta-judge score through a collaborative discussion among the agents.

\textbf{Hyperparameters:} 
The \textit{weight $w_{j}^{c}$} of each criterion in the rubric is determined based on its importance and can be dynamically adjusted for different scenarios. GPT-4 is tasked with assigning these weights during the rubric refinement process. All the criterion weights are shown in Table~\ref{tb:weight-c}.
The \textit{weight $w_{i}^{agent}$} is allocated based on the performance of agent $i$ on task. By default, all agents are treated equally, with $w_{i}^{agent}=1/{number(agents)}$. In some cases, if the agent $i$ demonstrates superior performance on a specific task, a higher weight could be assigned to $w_{i}^{agent}$.

\begin{table}[ht]
\centering
\setlength{\tabcolsep}{13pt}
\begin{tabular}{ll}
\toprule
\textbf{Criterion} &\textbf{$w_{j}^{c}$}  \\
\midrule
Accuracy of Judgment &  0.2\\
Logical Soundness  &  0.2\\
Completeness of Evaluation & 0.15 \\
Fairness & 0.1 \\
Relevance to Context & 0.15 \\
Clarity of Explanation & 0.1 \\
Impactfulness & 0.1 \\
\bottomrule
\end{tabular}
\caption{Hyperparameter for the weight of each criterion in the rubric. A higher weight means the criterion score has a larger proportion in the final score, making it more important.}
\label{tb:weight-c}
\end{table}

The selection threshold is set to a fixed value of 4.5 to ensure both high precision and robustness.

\subsection{Results}
\begin{table*}[ht]
\centering
\begin{tabular}{llcccccc}
\toprule
\textbf{Config} &\textbf{Agent} & \textbf{Knowledge} & \textbf{Reasoning} & \textbf{Math} & \textbf{Coding} & \textbf{Overall} \\
\midrule
Raw Judgments & \textbackslash & 62.34 & 52.04 & 78.57 & 59.52 & 61.71 \\
baseline & same as judge model&\textbf{73.68} &60.87 & 71.43& \underline{66.67}&  \underline{68.89}\\
\midrule
\textbf{SR} Selection&gpt-4o & 68.13 & 57.97 & 77.78 & 58.06 & 65.68\\
&gpt-4o-mini & \underline{71.62} & 62.50 & 76.32 & 60.00 & {68.65} \\
&claude & 65.71 & 63.49 & \textbf{79.17} & \textbf{70.50} & 67.47 \\
&llama & 66.14& 55.56 & 77.55 & 61.11 & 64.51 \\

\midrule
\textbf{LR} Selection&gpt-4o & 67.92& 58.33& 77.50& 52.17&65.24 \\
&gpt-4o-mini & 67.14 & \underline{65.91} & 75.68 & 60.00 & 67.84\\
&claude & {71.05} & \textbf{66.67} & \textbf{79.17} & 58.62 & \textbf{70.15} \\
&llama & 65.62& 54.22& 77.55& 60.53& 63.76 \\

\midrule
\textbf{BR} Selection&gpt-4o & 62.07& 54.76& {76.47}& {57.50}&61.88 \\
&gpt-4o-mini & 63.57& 54.76 & \underline{78.85}& 60.53& 63.38\\
&claude& 62.91& 51.58& 77.78& 58.54& 61.58 \\
&llama &63.70 &55.29 & 77.36 & 59.52 & 63.19 \\
\bottomrule
\end{tabular}
\caption{Meta-Judge Precision of different agents across various configurations. Raw judgments are from Arena-Hard Judge on JudgeBench using GPT-4o-mini. Configuration \textbf{LR} means Long Rubric, \textbf{SR} means Short Rubric, and \textbf{BR} means Binary Rubric.}
\label{tb:single-meta-judge-1}
\end{table*}

\begin{table*}[ht]
\centering
\begin{tabular}{llcccccc}
\toprule
\textbf{Config} &\textbf{Agent} & \textbf{Knowledge} & \textbf{Reasoning} & \textbf{Math} & \textbf{Coding} & \textbf{Overall} \\
\midrule
Raw Judgments & \textbackslash & 50.00& 53.06 & 55.36& 38.10 &50.29 \\
baseline & same as judge model& 50.99& 54.17& 59.62& 39.02& 51.76 \\
Selection &GPT-4o & \textbf{66.07}&\textbf{73.33} &\textbf{66.67} & \textbf{62.50} & \textbf{67.77}\\
\bottomrule
\end{tabular}
\caption{Meta-Judge Precision of different agents across various configurations. Raw judgments are from Arena-Hard Judge on JudgeBench using Llama-3.1-8B-Instruct. Selection uses a stronger meta-judging model with a rubric designed on the basis of Table~\ref{tb:single-meta-judge-1}.}
\label{tb:single-meta-judge-2}
\end{table*}

\subsubsection{Single agent precision comparison}
The current meta-judging method, which uses a one-sentence rubric to guide the judge model in evaluating its own judgments \cite{14-wu2024meta}, serves as the baseline. In comparison, various rubrics and meta-judging agents are evaluated to identify the optimal configuration for a single meta-judging model.

From Table~\ref{tb:single-meta-judge-1}, we observe that raw judgments generated by the GPT-4o-mini judge model exhibit low precision, with nearly half being incorrect. Meta-judge selection significantly improves precision in the selected judgments.
For the Knowledge task, using the same meta-judge model as the judging model with a concise baseline rubric achieves the highest precision, likely due to internal knowledge overlap.
For the Reasoning task, a longer, detailed rubric with a finer-grained scoring range improves precision, though overall performance remains lower due to task complexity.
For the Math task, adopting a more detailed and comprehensive rubric improves performance over the baseline.
For the Coding task, a simpler rubric is more effective, as criteria such as contextual relevance and fairness are less applicable to programming systems. Including irrelevant criteria can negatively impact meta-judging performance.

In conclusion, for knowledge tasks, using the same meta-judge model as the judging model is beneficial due to internal knowledge overlap, which enhances self-reflection. Complex reasoning tasks benefit from powerful large models and longer rubrics, while math tasks perform better with comprehensive rubrics. Although Claude, with a detailed rubric, achieves high meta-judge precision, its high variance suggests it is better suited for specific tasks. For diverse tasks, a simple baseline rubric shows more robust performance.

Table~\ref{tb:single-meta-judge-2} examines whether using a more powerful model (GPT-4o) for meta-judging, instead of relying on self-reflection, is more effective when the judge model (LLaMA-3.1-8B) has limited capability. The results show that adopting a stronger LLM for meta-judging, paired with a rubric designed based on the above conclusions, improves precision by approximately 16\%.

\subsubsection{Multi-agent precision comparison}
\begin{table*}[ht]
\centering
\begin{tabular}{lccccc}
\toprule
\textbf{Config}  & \textbf{Knowledge} & \textbf{Reasoning} & \textbf{Math} & \textbf{Coding} & \textbf{Overall} \\
\midrule
raw judgment collection & 62.34 & 52.04 & 78.57 & 59.52 & 61.71 \\
baseline & 73.68 &60.87 & 71.43& {66.67}&  {68.89}\\
\midrule
{Majority Voting} & \textbf{79.07} & \underline{69.23}& \underline{80.00} &\textbf{85.71} & \textbf{77.26} \\
{Weighted Average} & \underline{75.00} & \textbf{70.27}&78.72 & \textbf{85.71}& \underline{75.56} \\
{Panel Discussion} & 71.62 & 68.57& \textbf{80.43} & \underline{75.00} & 72.58\\
\bottomrule
\end{tabular}
\caption{Precision on Different multi-agent configurations. Raw judgments are from Arena-Hard Judge on JudgeBench using GPT-4o-mini. The multi-agent module includes the two best-performing models based on meta-judging: GPT-4o-mini and Claude.}
\label{tb:multi-agent}
\end{table*}

\begin{figure*}[htbp]
    \centering
    \includegraphics[width=0.8\linewidth]{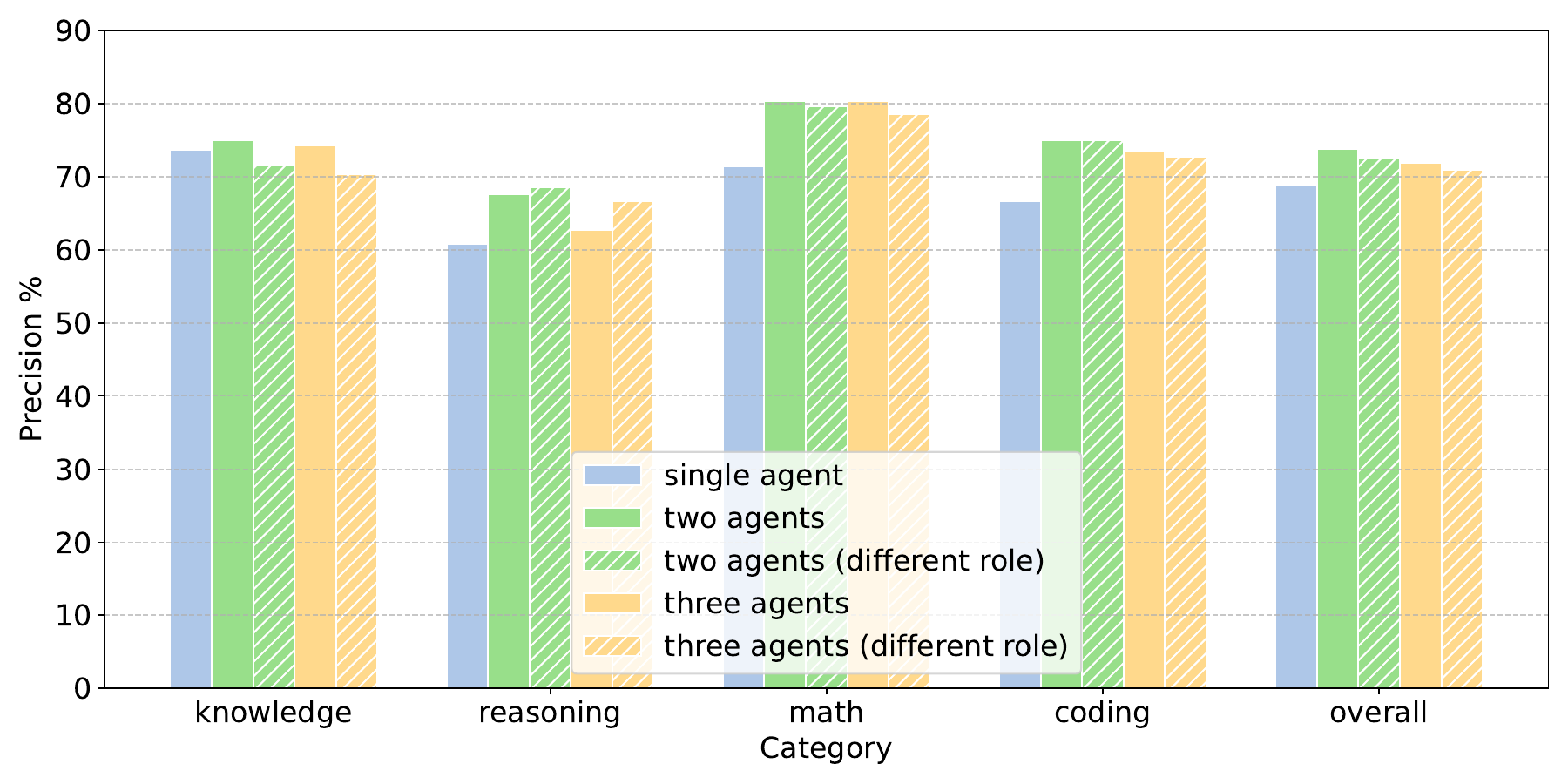} 
    \caption{Ablation study on the impact of the number and roles of agents across different tasks. The vertical axis represents the precision of the selected judgments by multi-agent meta-judging. The horizontal axis represents different tasks and the overall summary of all tasks.}
    \label{fig:ablation_study} 
\end{figure*}

Given the limitations of a single agent, where no single model can achieve optimal performance across all tasks, we propose utilizing a multi-agent module to address this challenge. Three multi-agent aggregation approaches, discussed in Section 3.2 are evaluated, and the results are presented in Table~\ref{tb:multi-agent}.

From Table~\ref{tb:multi-agent}, we can see that the majority voting strategy significantly improves meta-judge selection precision in knowledge and coding tasks, with gains of 6.61\% and 19.04\%, respectively. The weighted average strategy excels in reasoning tasks, achieving a 9.40\% improvement. Meanwhile, the panel discussion strategy performs particularly well in math tasks, yielding a 9.00\% improvement. 

These three strategies aggregate agents' capabilities from different perspectives. Majority voting and weighted averaging are forms of late aggregation; agents do not influence each other's generation. Majority voting selects a result agreed upon by more than half of the agents, while weighted averaging computes the final score by combining and averaging the opinions of all agents. In contrast, panel discussion represents early aggregation, where agents' opinions influence one another during the early stages, ultimately producing a final result. 
In Q\&A and multi-turn dialogue tasks, panel discussions consistently outperform other methods. However, when applied to meta-judging tasks, their performance fell short of expectations. 
This discrepancy could be attributed to the handling of challenging tasks, where maintaining diverse opinions among different agents helps improve the correctness of the final decision. However, during panel discussions, the opinions of different agents tend to converge over time, which is unfavourable for addressing extremely difficult problems.

\textbf{Computation Cost Analysis:} When each agent generates its meta-judge score independently, we only need to run each agent once. In this case, the total computation cost is given by $\sum_{i=1}^{N} cost_{i}$, where $N$ is the number of agents.
In contrast, if the agents engage in a discussion to exchange opinions, as illustrated in Figure~\ref{fig:discussion} (a) and (b), the computation cost increases. Specifically:
For the scenario depicted in Figure~\ref{fig:discussion} (a), the cost doubles to $2 *\sum_{i=1}^{N} cost_{i}$.
For the scenario depicted in Figure~\ref{fig:discussion} (b), an additional aggregation cost is incurred, yielding $2 *\sum_{i=1}^{N} cost_{i} + cost_{sum}$.

\subsubsection{Ablation study}

In this section, we evaluate the performance of various panel discussion structures to analyze the impact of the number of agents and distinct role assignments on meta-judging precision.

\paragraph{Number of Agents:}
Setting aside the summarization agent, we first evaluate the impact of the number of agents in the discussion panel, as shown in Fig~\ref{fig:ablation_study}. The results indicate that, across all tasks, a two-agent panel discussion structure performs best. However, when different roles such as expert, critic, and general public are assigned to each agent, the performance slightly decreases. Furthermore, increasing the number of agents to three does not lead to further improvement; instead, the performance shows a slight decline.
Possible explanations for this results could be: First, in the meta-judging task, each agent is assigned a role as a meta-judge, and introducing additional role descriptions could cause ambiguity in role definition. Second, the different roles in the meta-judging tasks do not result in significantly varied solutions to the problem. Third, as the number of panel agents increases, their cognitive errors may interact and amplify one another.

\paragraph{Implementation of the Summarization Agent:} 
With a fixed number of agents, we evaluate the impact of the summarization agent, as shown in Table~\ref{tb:abalation-2}. The results indicate that incorporating a summarization agent does not improve the performance of meta-judging. Instead, it slightly weakens the meta-judge capability of the multi-agent module. The result may be due to information overload, where the summarization agent must manage information for the question, responses, judgments, and all the meta-judgments from different agents.

\begin{table}[ht]
\centering
\begin{tabular}{cc}
\toprule
\textbf{Summarization}  &  \textbf{Overall} \\
\midrule
\ding{55} & 72.58\\
\ding{51} & 65.38\\
\bottomrule
\end{tabular}
\caption{Ablation study on the impact of the summarization agent. The multi-agent module consists of two agents. When the summarization agent is not utilized, majority voting is used to aggregate the scores and derive the final result.}
\label{tb:abalation-2}
\end{table}

\section{Discussion}
In our work, we put forward a multi-agents meta-judge selection framework. Through this framework, good judgments with high-quality explanations and correct decisive conclusions could be selected, and the precision of the selected judgment set could be improved.

Current works primarily focus on evaluating LLM judgments based on their alignment with human judgment, treating human judgment as the ground truth to optimize LLM performance. However, these studies overlook biases and mistakes in human judgment. Additionally, they do not propose methods for selecting judgments with correct conclusions and high-quality explanations. In contrast, our work demonstrates that LLMs can evaluate judgments without human intervention, accurately identifying good and bad judgments through the meta-judge score.

\textbf{Limitations:} Our experiment evaluates the performance of the meta-judge selection framework using a limited judgment set consisting of 350 judgments for response pairs generated by GPT-4o-mini. The limited dataset may restrict the generalizability of some of our experimental conclusions.

\textbf{Future Work:} The remarkable performance of LLMs as meta-judges shows their potential for leveraging meta-judge scores to construct preference datasets, facilitating the training of LLMs as judges and enhancing their performance in an unsupervised manner. As demonstrated in prior work \cite{14-wu2024meta} and \cite{trivedi2024self}, this approach can be further enhanced through our proposed meta-judge framework, which incorporates more sophisticated rubrics and aggregated results from multiple agent evaluations, thereby improving the reliability and robustness of the meta-judge process.

\section{Conclusion}
In this paper, we propose a meta-judge selection framework aimed at minimizing human intervention and fully leveraging LLMs for meta-judgment. The framework requires only a human-defined basic rubric at the initial stage, after which all subsequent steps are carried out autonomously by LLMs. Additionally, the source experimental dataset we adopted, JudgBench, reduces reliance on human annotations by utilizing algorithmically generated objective labels.
To further enhance evaluation accuracy, our framework incorporates a multi-agent module that effectively combines the strengths of multiple LLMs, overcoming the limitations of single-agent approaches. 

Experimental results on raw judgment sets derived from JudgBench demonstrate that LLMs can perform high-precision evaluations of their own judgments. The meta-judge scores produced by our framework reliably indicate judgment quality, enabling the precise identification of high-quality judgments with an appropriately chosen threshold.
This work lays a solid foundation for extending the capabilities of LLMs as judges to tackle superhuman-level tasks. By leveraging meta-judgment, we provide a promising pathway toward fully autonomous, scalable, and accurate evaluation systems driven by LLMs.



\section*{Impact Statement}
This paper presents work whose goal is to advance the field of Machine Learning. There are many potential societal consequences of our work, none which we feel must be specifically highlighted here.

\nocite{langley00}

\bibliography{example_paper}
\bibliographystyle{icml2025}

\newpage
\appendix
\onecolumn
\section{Detail of different rubrics}
In the single-agent precision comparison section, we analyzed the impact of four different rubric configurations. Section 3.1 provides an example of the 'logical soundness' criterion in a short rubric. Here, we offer more details regarding the baseline, long, and binary rubrics.
\begin{table}[h!]
    \centering
    \setlength{\tabcolsep}{8pt} 
    \renewcommand{\arraystretch}{1.5} 
    \begin{tabular}{p{0.95\columnwidth}}
        \toprule
       \rowcolor{gray!20} \textbf{Criterion:} \textit{Logical Soundness} \\
        \midrule 
         \textbf{Description:} \newline
         \textit{Assesses whether the judgment or decision follows a coherent and logical progression from the evidence or reasoning process. 
        A well-reasoned decision should clearly demonstrate how conclusions were drawn and avoid logical fallacies or contradictions. 
        This ensures the reasoning process is transparent and defensible.} \\
        \midrule
         \textbf{Score 1:} \textit{Decision-making process is illogical, lacking clear reasoning or consistency. 
         The conclusion appears arbitrary or disconnected from the supporting evidence.}\\ 
         \textbf{Score 2:} \textit{Decision-making process shows significant gaps or logical flaws, making it difficult to follow. 
            Reasoning is inconsistent, and critical errors undermine the validity of the conclusion.}\\ 
         \textbf{Score 3:} \textit{Decision-making process is moderately logical, but some inconsistencies or gaps weaken its coherence. 
            While the reasoning is partially sound, certain steps may appear unclear or unsupported.}\\ 
         \textbf{Score 4:} \textit{Decision-making process is mostly logical, with minor issues that do not undermine its overall integrity. 
            The reasoning is generally clear and follows a structured progression with only slight missteps.}\\ 
         \textbf{Score 5:} \textit{Decision-making process is entirely logical, with clear and consistent reasoning throughout. 
            Every step in the reasoning process is well-supported and leads naturally to the conclusion.}\\ 
        \bottomrule
    \end{tabular}
    \caption{Example of meta-judging long version rubric. This version offers a detailed three-sentence description for each criterion and a two-sentence explanation for each scoring range.}
    \label{tab:rubric-describe-1}
\end{table}

\begin{table}[h!]
    \centering
    \setlength{\tabcolsep}{8pt} 
    \renewcommand{\arraystretch}{1.5} 
    \begin{tabular}{p{0.95\columnwidth}}
        \toprule
       \rowcolor{gray!20} \textbf{Criterion:} \textit{Logical Soundness} \\
        \midrule 
         \textbf{Description:} 
         \textit{Measures if the decision-making process follows a clear and logical reasoning path.} \\
        \midrule
         \textbf{Prompt:} \textit{Finally, determine the correctness of the judgment and decision based on the above rubrics. Please respond strictly in the following format:  result: [correct/wrong], Explanation:}\\ 
        \bottomrule
    \end{tabular}
    \caption{Example of meta-judging binary version rubric. This version provides a brief description of each criterion and requires meta-judges to make a correct or incorrect determination based on all criteria collectively rather than evaluating them separately.}
    \label{tab:rubric-describe-2}
\end{table}

\begin{table}[h!]
    \centering
    \setlength{\tabcolsep}{8pt} 
    \renewcommand{\arraystretch}{1.5} 
    \begin{tabular}{p{0.95\columnwidth}}
        \toprule
       \rowcolor{gray!20} \textbf{Description:} 
         \textit{Explain which judgment is more accurate according to the original rubric and why.
        Combine judgment and decision to finally assign a score and consider factors such as adherence, accuracy, and consistency to the judgment instruction.} \\
        \midrule
         \textbf{Prompt:} \textit{Please respond strictly in the following format:  Score: [1-5], Explanation:}\\ 
        \bottomrule
    \end{tabular}
    \caption{Example of meta-judging baseline rubric. This baseline rubric is referred from the current paper regarding meta-judging\cite{14-wu2024meta}. This rubric is adapted from the current paper on meta-judging \cite{14-wu2024meta}. The baseline rubric does not include descriptions for individual factors, such as adherence and accuracy, and assigns an overall score encompassing all factors rather than separate scores for each.}
    \label{tab:rubric-describe-3}
\end{table}

\section{Panel discussion detail}
The panel discussion is conducted by recording the meta-judging history of each agent and providing it as a reference for other agents during meta-judging. The detailed prompt template and role description are shown in Table~\ref{table:template}. The role description is outlined in detail in Section B.1.

\begin{table}[h!]
\centering
\renewcommand{\arraystretch}{1.5} 
\setlength{\tabcolsep}{10pt} 
\rowcolors{1}{gray!20}{gray!20} 
\begin{tabular}{@{}p{4cm}p{12cm}@{}} 
\textbf{[Role description]} & \texttt{you are a ......} \\ 
\textbf{[Question]} & \texttt{Question from mmlu-pro.} \\ 
\textbf{[Answer A]} & \texttt{response A} \\ 
\textbf{[Answer B]} & \texttt{response B} \\ 
\textbf{[Judgment]} & \texttt{Both assistants provided incomplete solutions, but Assistant A's approach is closer to the correct logic. Assistant B's condition is fundamentally incorrect.}\\ 
\textbf{[Decision]} & \texttt{[A > B]}\\ 
\textbf{[meta-judgment history]} & \texttt{Criterion: Accuracy of Judgment \newline
Score: 5  \newline
Explanation: The judgment accurately interprets the problem and correctly identifies Assistant A has more correct logic.}\\ 
\textbf{[System]} & 
While scoring, refer to the meta-judging results from other agents. Identify the agreed-upon opinions and analyze any differing viewpoints.\\ 
\end{tabular}
\caption{The prompt template for a panel discussion with different role descriptions.}
\label{table:template}
\end{table}

\subsection{Role description for agents}

\textbf{General Public}

\textit{You are a general public meta-judge assistant designed to ensure fairness in evaluating the quality of the judgment and decision made by a judge assistant.}

 \textbf{Expert}
 
\textit{You are an expert meta-judge assistant with advanced expertise in evaluating the quality of the judgment and decision made by a judge assistant.}
 
 \textbf{Critic}

\textit{You are a critic meta-judge assistant tasked with providing critical analysis in evaluating the quality of the judgement and decision made by a judge assistant.}
 
 \textbf{Summarization}

\textit{You are a meta-judge coordinator assistant in aggregating the meta-judgments from other agents.}


\end{document}